\documentclass[letterpaper, 10 pt, conference]{ieeeconf}  % Comment this line out if you need a4paper

\IEEEoverridecommandlockouts                              % This command is only needed if 
\usepackage{times}
\usepackage{epigraph}

\usepackage{graphics,graphicx,caption,float,subcaption,booktabs,xcolor,multirow,multicol,array,color,ifthen,tabu,colortbl,dblfloatfix,url,xparse,mathtools,patchcmd,algorithm,algorithmic,amssymb,xspace,nicefrac,microtype,amsmath,amsfonts,bm,ragged2e,tikz,stackengine,etoolbox,xpatch,enumerate,xstring,setspace,tabularx,makecell,changepage,cuted,titlesec,wrapfig,tcolorbox, sidecap, xspace, bbm, algorithm, listings,hyperref}

\usepackage{etoolbox}
\makeatletter
\AfterEndEnvironment{algorithm}{\let\@algcomment\relax}
\AtEndEnvironment{algorithm}{\kern2pt\hrule\relax\vskip3pt\@algcomment}
\let\@algcomment\relax
\newcommand\algcomment[1]{\def\@algcomment{\footnotesize#1}}
\renewcommand\fs@ruled{\def\@fs@cfont{\bfseries}\let\@fs@capt\floatc@ruled
  \def\@fs@pre{\hrule height.8pt depth0pt \kern2pt}%
  \def\@fs@post{}%
  \def\@fs@mid{\kern2pt\hrule\kern2pt}%
  \let\@fs@iftopcapt\iftrue}
\makeatother
\newcommand{\AG}[1]{}
\newcommand{\todo}[1]{}
\newcommand{\VK}[1]{}
\newcommand{\pg}{pre-grasp\xspace}
\newcommand{\pgs}{pre-grasps\xspace}
\newcommand{\fname}{PGDM\xspace}
\newcommand{\suitename}{TCDM\xspace}
\newcommand{\website}{\href{https://pregrasps.github.io/}{https://pregrasps.github.io/}}

\begin{document}

% \title{PreGrasp is all you need:\\ a universal prior for Dexterous Manipulation}
\title{Learning Dexterous Manipulation from\\ Exemplar Object Trajectories and Pre-Grasps}
% Suggestions ===
% Learning Dexterous Manipulation from exemplar Object Behavior and Pre-Grasps
% Learning Dexterous Manipulation from specified Object Behavior and Pre-Grasps
% Learning Dexterous Manipulation for specified Object Behavior using Pre-Grasps
% Leveraging Pre-Grasps to Learn Dexterous Manipulation for exemplar Object Behaviors

% You will get a Paper-ID when submitting a pdf file to the conference system
\author{\textbf{Sudeep Dasari$^{1}$, Abhinav Gupta$^{1}$, Vikash Kumar$^{2}$}\\
 \textit{Carnegie Mellon University$^{1}$, Meta AI Research$^{2}$}
\thanks{An abridged version of this paper was presented in IEEE International Conference on Robotics and Automation (ICRA) 2023. This work was completed while the first author was an intern at Meta AI. Direct correspondence to:
        {\tt\small sdasari@cs.cmu.edu}}%
}

%\author{\authorblockN{Michael Shell}
%\authorblockA{School of Electrical and\\Computer Engineering\\
%Georgia Institute of Technology\\
%Atlanta, Georgia 30332--0250\\
%Email: mshell@ece.gatech.edu}
%\and
%\authorblockA{Twentieth Century Fox\\
%\and
%\authorblockN{James Kirk\\ and Montgomery Scott}
%\authorblockA{Starfleet Academy\\
%San Francisco, California 96678-2391\\

% avoiding spaces at the end of the author lines is not a problem with
% conference papers because we don't use \thanks or \IEEEmembership

%Homer Simpson\authorrefmark{2},
%Montgomery Scott\authorrefmark{3} and
%Eldon Tyrell\authorrefmark{4}}
%Telephone: (800) 555--1212, Fax: (888) 555--1212}
%\authorblockA{\authorrefmark{4}Tyrell Inc., 123 Replicant Street, Los Angeles, California 90210--4321}}

\maketitle

\begin{abstract}

Learning diverse dexterous manipulation behaviors with assorted objects remains an open grand challenge. 
While policy learning methods offer a powerful avenue to attack this problem, they require extensive per-task engineering and algorithmic tuning. This paper seeks to escape these constraints, by developing a \textbf{P}re-\textbf{G}rasp informed \textbf{D}exterous \textbf{M}anipulation (\textbf{\fname}) framework that generates diverse dexterous manipulation behaviors, without any task-specific reasoning or hyper-parameter tuning. At the core of \fname is a well known robotics construct, \textit{\pgs} (i.e. the hand-pose preparing for object interaction). This simple primitive is enough to induce efficient exploration strategies for acquiring complex dexterous manipulation behaviors. To exhaustively verify these claims, we introduce \textbf{\suitename}, a benchmark of 50 diverse manipulation tasks defined over multiple objects and dexterous manipulators. Tasks for \suitename are defined automatically using \textit{exemplar object trajectories} from various sources (animators, human behaviors, etc.), without any per-task engineering and/or supervision. Our experiments validate 
that \fname's exploration strategy, induced by a surprisingly simple ingredient (single \pg pose), matches the performance of prior methods, which require expensive per-task feature/reward engineering, expert supervision, and hyper-parameter tuning. For animated visualizations, trained policies, and project code, please refer to \website.

\end{abstract}

\section{Introduction} 
\label{sec:intro}

\begin{figure}
 \vspace{-0.2in}
    \begin{minipage}{1\textwidth}
    \centering 
    \includegraphics[width=\linewidth]{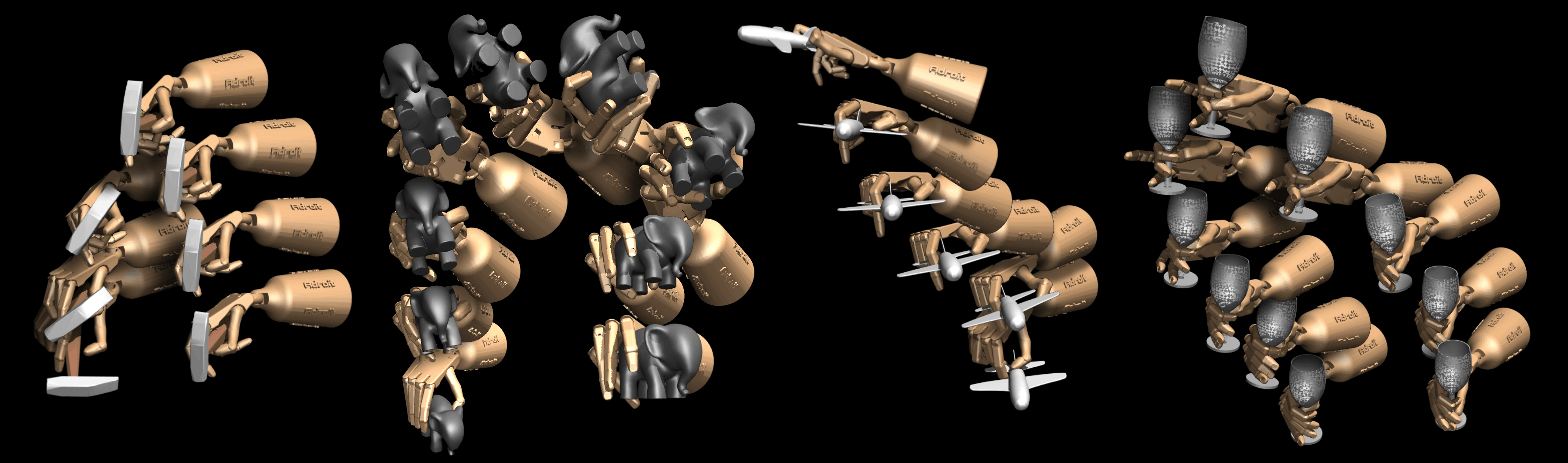} 
        \caption{Dexterous behaviors learned by \fname. Our experiments span diverse objects and tasks.}
    \label{fig:teaser}
    \end{minipage}
    \vspace{-0.3in}
\end{figure}

Dexterous manipulation tasks -- loosely defined as controlling a robot hand to effectively re-arrange its own environment~\cite{mason2018toward} -- were often solved by designing controllers to realize a sequence of stable object transitions. These approaches were limited to narrowly-scoped scenarios, as they required experts to carefully reason about hand-object contact behavior on a per-task basis: e.g. in terms of geometry~\cite{miller2004graspit,miller2003automatic,roa2007geometrical}, and/or force closures~\cite{howard1996stability,bicchi1995closure,liu2004complete,saut2005online}. 

\vspace{5cm}
.

\vspace{4.75cm}
As a result, the field has trended towards robot learning paradigms~\cite{lillicrap2015continuous,arulkumaran2017deep, hussein2017imitation}, which seek to leverage data-driven exploration to \textit{automatically} acquire dexterous behaviors~\cite{kumar2016optimal, kumar2016learning, nagabandi2020deep}. However, learning algorithms rely on naive search strategies~\cite{mania2018simple} that struggle with dexterous manipulation, due to high-dimensional search spaces (e.g. ShadowHand~\cite{Kumar2016thesis} has 30-DoF) and small error-margins. In practice, this ``deep exploration challenge" is addressed using a wealth of task-specific supervision and engineering. While effective, this approach takes us back to square one (heavy reliance on expert knowledge). 

This paper seeks to find a general strategy that can overcome the aforementioned exploration challenge with minimal assumptions. Instead of viewing this issue from a purely algorithmic perspective, we demonstrate that ``pre-grasp" states --  i.e. a classical robotics construct~\cite{kang1994determination} that denotes states where the robot is poised to initiate object interaction  -- can act as a general supervisory structure to relax exploration challenges in learning dexterous manipulation behaviors. Specifically, \pgs act as a ``pre-condition"  that can enable robots to efficiently and safely explore object contacts and intermittent interaction dynamics, without requiring exquisite optimization techniques. In addition, \pgs are practical: they are easy to specify (e.g. via human annotation), realize (movement in free space), and unlike grasps (see \ref{appendix:pg_vsgrasp} for details), do not involve hard to sense object surface or inertial details. Thus, we propose a \textbf{P}re-\textbf{G}rasp informed \textbf{D}exterous \textbf{M}anipulation (\textbf{\fname}) framework that embeds \pg poses (as exploration primitives) into existing learning pipelines, to synthesize behaviors without requiring task engineering or hyper-parameter tuning. While the connection between \pgs and manipulation has been studied before~\cite{kappler2010representation,kappler2012templates,baek2021pre,christen2022d}, to the best of our knowledge, we are the first to analyze \pg's effectiveness in a multi-task learning paradigm.

To demonstrate \fname's versatility, it is important to test across as many scenarios as possible. However, existing dexterous manipulation benchmarks are shallow and packed with expert knowledge (e.g. favourable tasks with heavily engineered details -- initialization, input, reward, etc.). Thus, we developed a \textbf{T}rajectory \textbf{C}onditioned \textbf{D}exterous \textbf{M}anipulation benchmark, \textbf{\suitename}. As the name suggests, \suitename tasks are automatically constructed from diverse \textit{exemplar object trajectories} (sourced from human behaviors, animations, etc.), which prevents expert designers from injecting task-specific supervision. Indeed, every part of the task setup (e.g. formulation, reward/termination functions, hyper-parameters, etc.) is kept constant across \suitename, except for the exemplar trajectory itself. \suitename's diverse tasks span: 3 robotic hands, 30+ standardized objects~\cite{brahmbhatt2019contactdb,calli2015ycb}, and behaviors ranging from fixed goal-reaching (e.g. relocation~\cite{rajeswaran2017learning,qin2021dexmv}) to cyclic, dynamic skills (e.g. hammering, bottle shaking, etc.).

This work develops \fname, a simple exploration framework for dexterous manipulation, and validates it across the diverse tasks in \suitename. Our contributions include:
\textbf{(1)} identifying \pgs as a key ingredient to guide successful exploration in dexterous manipulation, and embedding them into existing behavior synthesis pipelines. 
\textbf{(2)} Next, our \fname framework achieves SOTA results on a diverse suite of dexterous manipulation tasks, while using significantly less supervision (e.g. single frame vs full hand trajectory) than representative baseline methods. 
\textbf{(3)} In addition, we find which \pg properties (e.g. proximity, finger pose) are important for successful behavior learning. 
\textbf{(4)} Finally, we commit to open-sourcing the \suitename benchmark, \fname's code-base, and all experimental artifacts (e.g. trained policies) for the community's benefit.

% \begin{figure}[!t]
%   \centering
%   \includegraphics[width=\linewidth]{figures/teaser.png} 
%   \caption{Dexterous behaviors learned by \fname. Our experiments span diverse objects and tasks.}

%   \vspace{-0.15in}
% \end{figure}

\section{Related Work} 
\label{sec:related}

Prior robot learning approaches achieved impressive results on various dexterous tasks~\cite{chen2022system,huang2021generalization,gupta2021reset,rajeswaran2017learning,qin2021dexmv,mandikal2020dexterous,nagabandi2020deep,ahn2020robel,akkaya2019solving}. 
But while learning approaches strive to be automatic, prior work requires a wealth of expertise for successful deployment. We now classify these approaches (and others), by the supervision strategies required to make them work in practice.  

\paragraph{Task-engineering} A popular solution is to carefully design environments, tasks, and learning curriculums that structure the robot's exploration. This can be accomplished by: extensive reward shaping~\cite{nagabandi2020deep,akkaya2019solving} to reduce noise in optimization; action space constraints~\cite{rajeswaran2017learning} to prevent degenerate solutions; decomposing tasks into sub-skills~\cite{karpathy2012curriculum,portelas2020automatic}; changing environment physics so the policy can develop its skill over the course of training~\cite{chen2022system,akkaya2019solving}; and cleverly initializing the policy so it can learn to pass through challenging bottlenecks in the state space~\cite{peng2018deepmimic,florensa2017reverse}. These strategies require \textit{weeks} of expert trial and error to make a single-task solution, and often rely on unrealistic assumptions (e.g. changing gravity, reset robot in mid-air, etc.). In contrast, \fname avoids these issues by using a simple \pg based exploration primitive to accelerate learning, and needs no task knowledge.

\paragraph{Expert Data} Another common strategy is to initialize exploration strategies with expert data -- ranging from trajectories collected by human demonstrators~\cite{rajeswaran2017learning,handa2020dexpilot,qin2021dexmv} to affordances~\cite{mandikal2020dexterous} mined from human contact data~\cite{brahmbhatt2019contactdb,taheri2020grab}. However, this data rarely generalizes between settings (i.e. trajectories are robot and task specific), and collecting it is expensive, since it requires special purpose experimental setups~\cite{handa2020dexpilot,brahmbhatt2019contactdb}. More fundamentally, it is unclear if/when adding additional data can yield performance benefits. Our investigation addresses these issues, by outlining how simple data sources (\pgs and object trajectories) can accelerate policy learning, while being easy to acquire~\cite{rong2021frankmocap,ye2021shelf,cao2021reconstructing,liu2021semi}.

\begin{figure*}[!t]
    \centering
    \includegraphics[width=\linewidth]{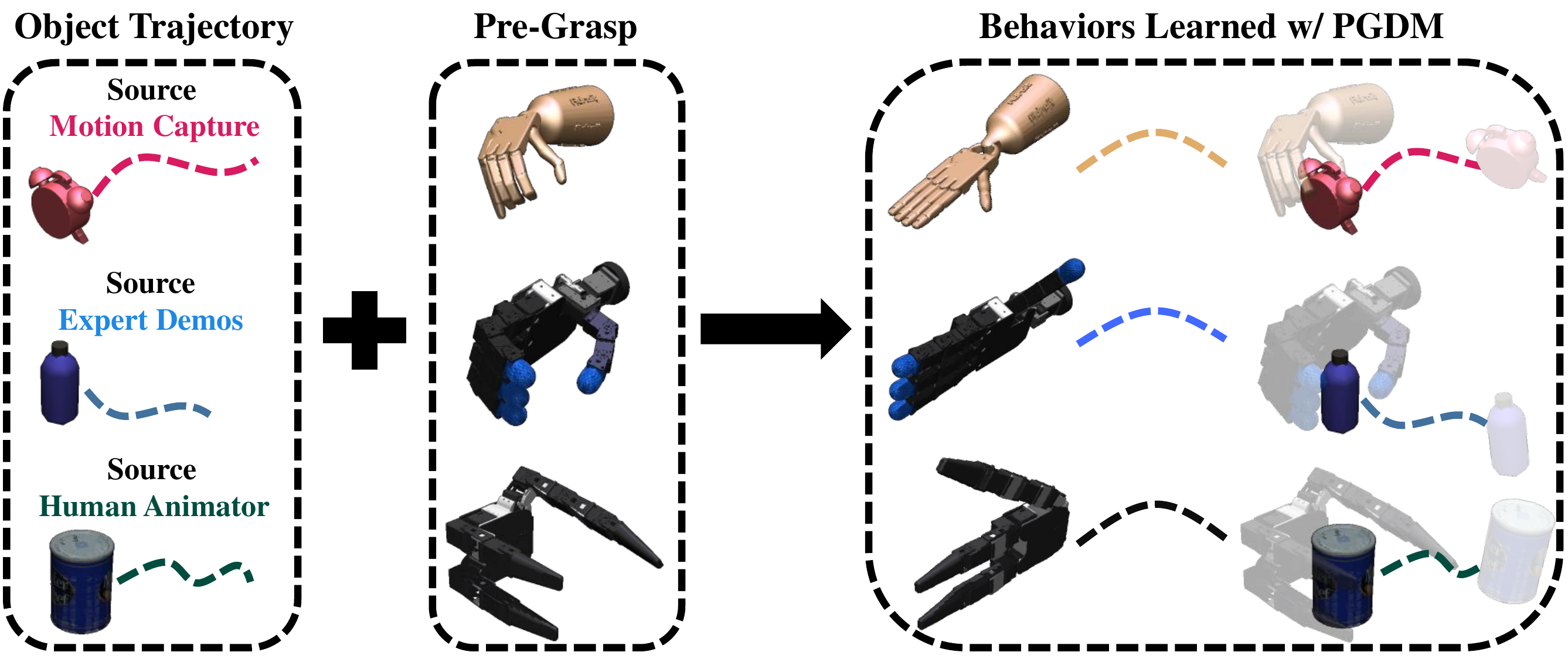} 
    \caption{Our method uses simple ingredients to learn dexterous behaviors in diverse scenarios. The \fname framework leverages \textit{\pg} states as general exploration primitives, to solve tasks automatically defined using \textit{object trajectories}, without any per-task tuning. Both of these ingredients can be easily mined (e.g. from MoCap data) or hand-annotated (e.g. by human animators).} 
    \label{fig:overview}
    \vspace{-0.1in}
\end{figure*}

\section{Methods}
\label{sec:method}

How can a single method learn a diverse range of dexterous manipulation behaviors? We argue that a simple, data-driven solution with minimal hyper-parameters offers the best chance. In this spirit, Sec.~\ref{sec:method_formulation} presents a general task formulation, which parameterizes diverse dexterous behaviors using \textit{exemplar object trajectories}, and Sec.~\ref{sec:method_pg} introduces \fname, a framework for accelerating policy learning using \pg states. An overview of our approach is shown in Fig.~\ref{fig:overview}.

\subsection{Task Formulation}
\label{sec:method_formulation}

Let's begin by formalizing the definitions for robotic tasks and environments. We adopt the finite Markov Decision Process (MDP) formulation~\cite{sutton2018reinforcement}. At each time-step the agent observes states ($s_t \in \mathcal{S}$) and goals ($g_t \in \mathcal{S}$), and executes an action ($a_t \in \mathcal{A}$). The next state evolves according to stochastic dynamics ($s_{t+1} \sim P(\cdot | s_t, a_t)$). The agent collects trajectory rollouts within the MDP ($\tau = [s_0, a_0, s_1, \dots, s_n]$), starting from an initial state $s_0 \sim P(s_0)$. Desired behavior is specified by adding \textit{time-varying} goal variables ($G = [g_1, \dots, g_T]$) that are used to condition both the reward function $R(s_t, a_t, g_t)$ (optimized by agent) and the termination condition $T(s_t, g_t)$ (early-stops failed episodes). To preserve the Markov property, the current time-step $t$ must be appended to $s_t$, since $g_t = G(t)$. Note that this is a super-set of the more standard static-goal conditioned MDP: it allows us to specify time-varying behaviors, and we can recover the standard formulation by setting $g_t = g \hspace{1mm} \forall  \hspace{1mm} t$. Given a discount factor $\gamma$, the learning objective is to find a policy $a_t \sim \pi( \cdot | s_t)$ that maximizes: $\textbf{max}_{\pi} \hspace{2mm} J(\pi) = \hspace{2mm} \mathbb{E}_{\tau \sim \pi} [\Sigma_{t=0}^\infty \hspace{0.5mm} \gamma^t R(s_t, a_t, g_t)]$. 

\paragraph{Parameterizing Task MDPs} We now describe how to create task MDPs from exemplar object trajectories -- i.e. $X = [x_1, \dots, x_T]$, where each $x_i = [x_i^{(p)}, x_i^{(o)}]$ is an object pose (position and orientation). Object trajectories are used as goal variables (i.e. $G = X$), which in turn parameterize a pre-defined reward function $R$ and termination condition $T$. Specifically: \textbf{(1)} goal variables are set to match the desired object pose at each time-step $g_t = x_t$; \textbf{(2)} the reward function encourages matching the exemplar trajectory -- $R(\hat{x}_t, x_t) := \lambda_1 exp\{-\alpha || x_t^{(p)} - \hat{x}_t^{(p)} ||_2 - \beta |\angle (x_t^{(o)}, \hat{x}_t^{(o)})|\} + \lambda_2 \mathbbm{1}\{lifted\}$, where $\hat{x}_t = \hat{x}(s_t)$ is the real object pose in state $s_t$, $\angle$ is the Quaternion angle between the two orientations, and $\mathbbm{1}\{lifted\} = x_t^{(z)} > \zeta \textbf{ and } \hat{x}_t^{(z)} > \zeta$ encourages stable object lifting; \textbf{(3)} episodes are terminated when the object is too far from the goal $T(x_t, \hat{x}_t) := ||x_t - \hat{x}_t ||_2 > \omega$. All hyper-parameters for these function are reported in App.~\ref{appendix:hyper_formulation}. This formulation encourages the robot to produce behaviors that match the given template object trajectory. In practice, it allows us to specify diverse tasks -- including dynamic, cyclic behaviors that eluded past work (e.g. hammering) -- simply by supplying an appropriate object trajectory. One can even recover standard static-goal conditioned behaviors (e.g. lifting), by setting a fixed goal pose for the whole trajectory. Note that all this is possible without any per-task engineering (e.g. knob-turning bonus~\cite{rajeswaran2017learning}, etc.).

\subsection{\fname: Accelerating Exploration w/ Pre-Grasps}
\label{sec:method_pg}

Our attention turns to creating a general exploration primitive that can accelerate learning across a wide range of tasks. Note that all dexterous tasks begin with the hand gaining proximity to the target object, before transitioning into general manipulation. Thus, it's natural to decompose dexterous tasks into a ``reaching stage" and a ``manipulation stage," and use different strategies to solve each. But in the first stage, what state should the robot reach for?  We argue that \pg states (i.e. hand pose directly preceding contact) provide the answer. Pre-grasps favourably position the robot relative to a target object, so that it can quickly learn the intermittent contacts behaviors required for dexterous manipulation. For example, the \pgs shown in Fig.~\ref{fig:overview} position the robot hand near the target object and \textit{around} functional parts (e.g. fingers wrapped around handle), which allows the robot to easily gain control. As an added bonus \pgs require minimal assumptions: they can be cheaply annotated by human labellers or mined from human behavior data~\cite{ye2021shelf,cao2021reconstructing}, and can be easily reached by robots (e.g. w/ free-space planner~\cite{karaman2011anytime}). The key insight of \fname is to exploit these favorable properties, by moving robot hands to the \pg state before beginning the learning process. From the learning agent's (i.e. $\pi$) perspective, this is equivalent to modifying $P_0$ to maximally reduce exploration complexity, while still making minimal assumptions in practice. For additional \pg examples and a more extensive definition, please refer to App.~\ref{appendix:pg}.

\section{Experimental Setup}
\label{sec:implement}

The following sections describe how our task formulation is used to create \suitename (see Sec.~\ref{sec:implement_suite}), alongside our implementation of \fname (see Sec.~\ref{sec:implement_fname}). We stress that these decisions were made for the sake of consistent experiments, and are not inherent to our framework. 

\subsection{Introducing \suitename}
\label{sec:implement_suite}

Our task formulation (see Sec.~\ref{sec:method_formulation}) acts as a recipe for converting exemplar object trajectories into dexterous manipulation tasks. We use this to define a set of 50 tasks, which span: 34 different objects; 3 distinct robotic hand platforms; and unique object trajectories mined from three sources -- motion capture data-sets, human animated trajectories, and expert policy behaviors (see Figs.~\ref{fig:teaser},~\ref{fig:overview}). The \pgs for each task come from one of four sources: (1) human MoCap recordings~\cite{taheri2020grab} transferred to robot via IK, (2) expert \pgs extracted from Tele-Op data~\cite{rajeswaran2017learning}, (3) manually labeled \pgs, and (4) learned \pgs generated by an object mesh conditioned grasp predictor~\cite{taheri2020grab}. Further details on the task creation process and a full table of all tasks are presented in App.~\ref{appendix:hyper_suite}.

To make an investigation of this scale reproducible, the tasks are simulated (using MuJoCo~\cite{todorov2012mujoco}) and compiled into a benchmark, named \suitename-50. Note that a subset of 30 tasks (named \suitename-30) contain additional supervision, in the form of expert hand trajectories and grasping data. While not useful for our formulation, this data is required for the baselines. Thus, some of our experiments are run on the abridged \suitename-30 for fair comparison. More details are provided in App.~\ref{appendix:hyper_sim}.

\paragraph{Success Metrics:} Before continuing, let's discuss quantitative metrics for judging performance on \suitename tasks. Put simply, a ``good" policy is one that stably controls the object and matches the exemplar trajectory. We define (using the constants from Sec.~\ref{sec:method_formulation}) three simple metrics that capture both these properties. The \textit{COM Error} metric -- $E(\hat{X}) = \frac{1}{T} \Sigma_{t=0}^T || x_t^{(p)} - \hat{x}_t^{(p)} ||_2$ -- calculates Euclidean error (in meters) between the object's COM position, and the desired position from the exemplar trajectory. Similarly, the \textit{Ori Error} metric -- $E(\hat{X}) = \frac{1}{T} \Sigma_{t=0}^T \angle (x_t^{(o)} - \hat{x}_t^{(o)})$ -- is defined as the angle (in radians) between the achieved object orientation and desired orientation. In addition, the \textit{success} metric -- $S(\hat{X}) = \frac{1}{T} \Sigma_{t=0}^T \mathbbm{1}\{|| x_t^{(p)} - \hat{x}_t^{(p)} ||_2 < \epsilon \}$ reports the fraction of time-steps where COM error is below a $\epsilon=1$cm threshold. In practice, we've noticed that that humans easily perceive roll-outs that score from $60-80\%$ as ``successful." 

\subsection{Implementing \fname}
\label{sec:implement_fname}

Finally, let's discuss our implementation for the two stage task decomposition proposed by \fname (see Sec.~\ref{sec:method_pg}). Given a task's initial state distribution $P(s_0)$ (e.g. hand at reset position and object on table), we load an appropriate \pg state $s_{pg}$, and solve for a policy (using a scene-agnostic trajectory optimizer~\cite{Lowrey-ICLR-19}) that moves the robot to $s_{pg}$. At this point, our system transitions to learning an agent within the MDP (i.e. maximize $J$) as normal. Specifically, we utilize the PPO~\cite{schulman2017proximal} algorithm, to learn the dexterous behavior. Note that (except for \pg) the entire system remains fixed across different tasks. All relevant hyper-parameters and pseudo-code are presented in App.~\ref{appendix:hyper_framework}.
\section{Experiments}
\label{sec:experiments}

These experiments seek to validate both our trajectory centric task formulation (i.e. \suitename) and our \pg based exploration primitive (i.e. \fname). Specifically, we pose the following questions: \textbf{(Q1)} Can our methodology learn a broad and diverse range of dexterous manipulation skills? \textbf{(Q2)} Are we able to match the performance of baselines methods that leverage task specific reasoning (demonstrations, curriculum, etc.)? \textbf{(Q3)} What attributes of \pgs make them useful exploration primitive? \textbf{(Q4)} And finally, how accurately do our simulated results match real world behavior?

\subsection{Learning Behaviors w/ Pre-Grasps and \fname Tasks}
\label{sec:exp_validate}

\begin{table}
        \centering
        \vspace{-0.15in}
        \resizebox{\linewidth}{!}{%
            \begin{tabular}{lc|cccc}
                \toprule
                % \rowcolor{lightgray}
                 & \textbf{All Trials} & MoCap & Tele-Op & Labeled & Learned\\
                 
                 \midrule
                 \textit{Success} & $74.5\%$ & $75.0\%$ & $84.5\%$ & $69.2\%$ & $90.4\%$\\
                 
                 \textit{COM Error (m)} & $5.23$e-3 & $4.45$e-3 & $1.12$e-3 & $7.76$e-3 & $1.55$e-3\\

                 \textit{Ori Error (rad)} & 0.33 & 0.32 & 0.059 & 0.42 & 0.18 \\

                 \textit{\# of Tasks} & 50 & 37 & 3 & 7 & 3\\
                \bottomrule
            \end{tabular}
        }
        \caption{ Error and success metrics (averaged across all tasks w/ 3 seeds per-task) at the end of RL training (50M samples), and broken down by pre-grasp source. Note how \fname achieves high performance using diverse pre-grasp sources.}
        \vspace{-0.1in}
        \label{tab:validate}
    
\end{table}

To verify our methods' viability, we deploy the \fname policy learning scheme on the \textit{entire} \suitename benchmark. Recall, no task specific tuning is allowed for \textit{any} component in our setup. Since RL algorithms often display significant run-to-run variance, we run this experiment using 3 random seeds. Error and success metrics at the end of training (broken down by pre-grasp source and averaged across tasks) are shown in Table~\ref{tab:validate}. The behavior policies learned with \fname achieve a tracking error of $5.23$e-3, success rate of $74.5\%$, and low run-to-run variance, despite the breadth and diversity of \suitename tasks. Note that \fname can learn effective policies using any of our 4 \pg sources (MoCap, Expert Tele-Op, Human Labeled, and Learned). In particular, the successful experiments w/ learned \pgs suggest further avenues for scaling our results. Even though \fname uses no hand supervision outside of \pgs, we note that the final policies often produce smooth motions and realistic finger behavior. This defies conventional wisdom in the field that suggests human supervision is critical for ``normal" behaviors~\cite{rajeswaran2017learning,mandikal2020dexterous,qin2021dexmv}. That being said, multiple imperfections (e.g. large forces) remain in our policies, which leaves room for further improvement. Readers are encouraged to view the supplementary video\footnote{\website}, to understand the learned qualitative behaviors. For additional visualizations, learning curves, and a more thorough breakdown of individual tasks please refer to App.~\ref{appendix:all_tasks}. 

\subsection{Baselines: Is Additional Supervision Needed for Exploration?}
\label{sec:exp_baselines}

\begin{figure*}[t]
    \centering
    \begin{minipage}{\linewidth}
        \includegraphics[width=0.49\linewidth]{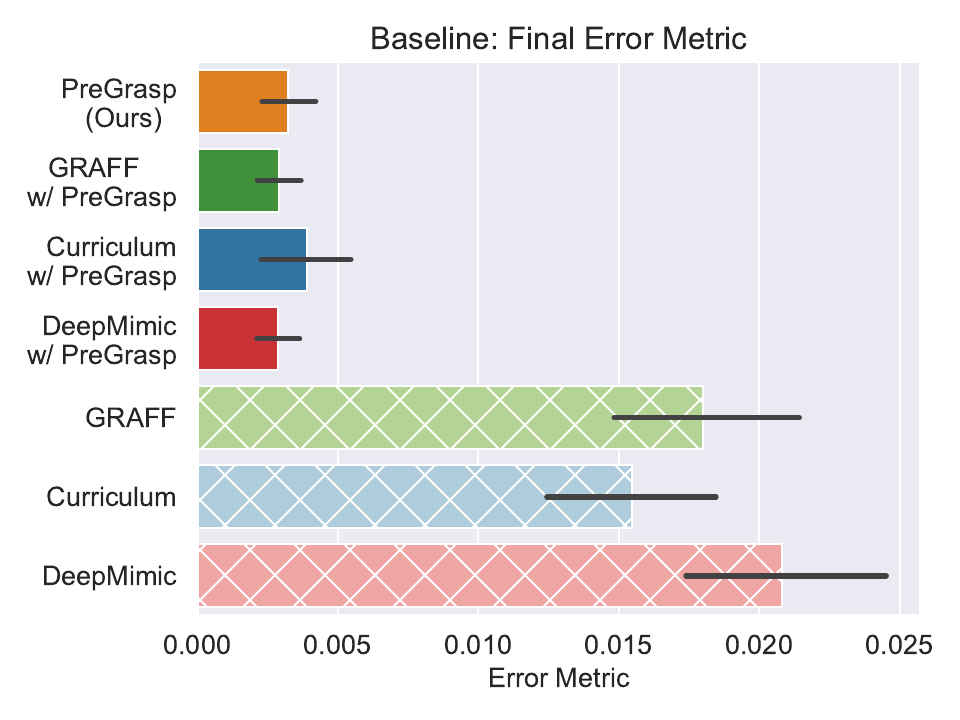}
        \includegraphics[width=0.49\linewidth]{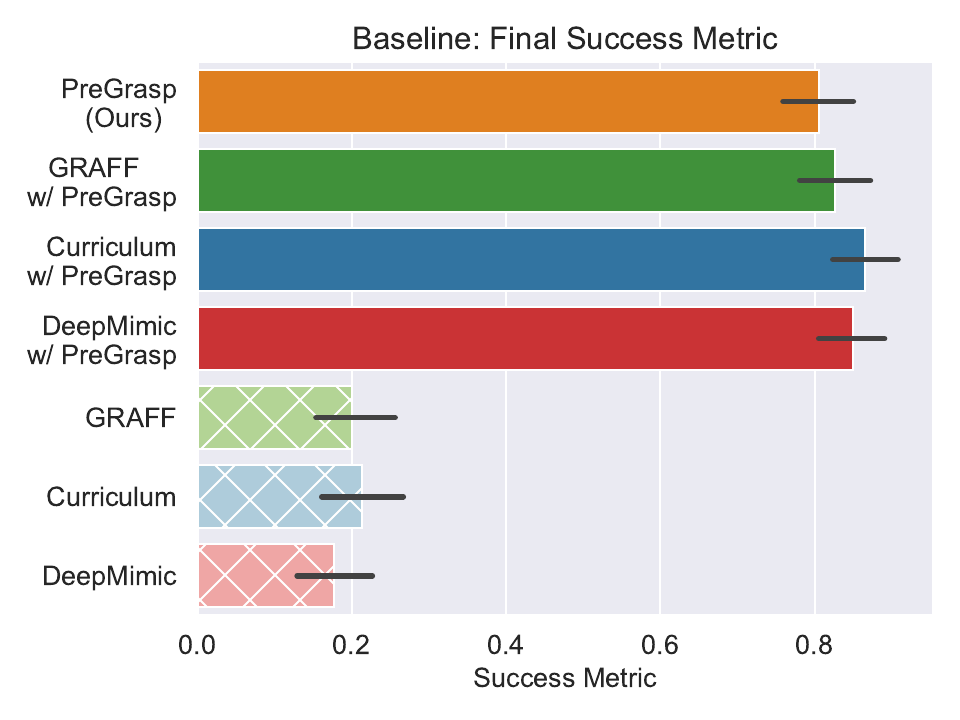}
        \caption{Average Success and Error metrics at the end of training for both the \fname-only method and 6 baselines (3 methods, w/ and w/out \fname). Note how methods using \fname strongly outperform those that don't, and how adding additional supervision does not improve performance.}
        \label{fig:baselinebar}
    \end{minipage}
    
    \begin{minipage}{\linewidth}
        \vspace{0.1in}
        \includegraphics[width=0.49\linewidth]{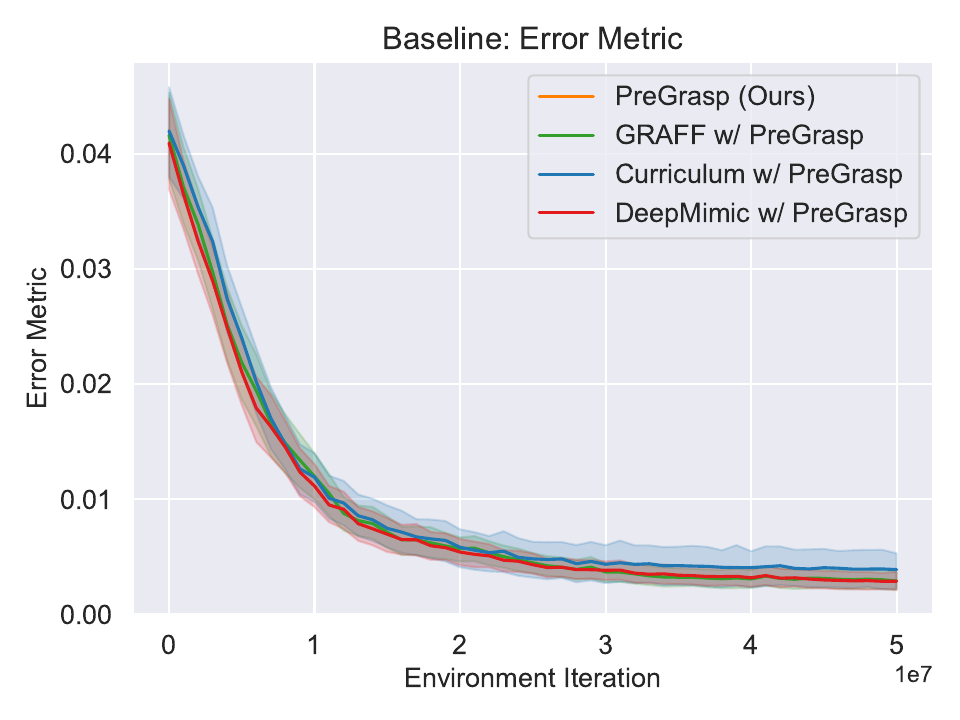}
        \includegraphics[width=0.49\linewidth]{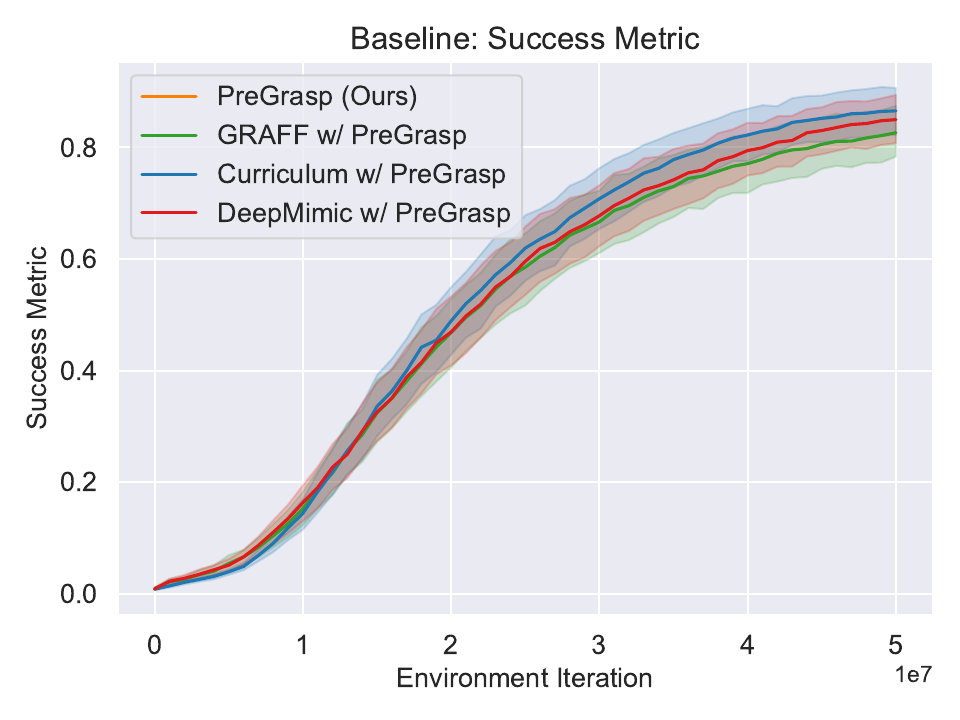}
        % \vspace{-0.12in}
    
        \caption{Learning curves (of Success/Error metric) comparing \fname-only against baselines initialized w/ \fname. Note how adding supervision to \pgs does not improve performance at any stage of learning.}
        \label{fig:baselinecurve}
    \end{minipage}
    
    % \caption{Learning curves displaying our Error and Success metrics averaged across 30 tasks from \suitename (each w/ single seed). The top plots compare \pgs v.s. our baselines + \pg, whereas the bottom plots compare \pgs vs just the baselines. Note how our method performs comparably against DeepMimic+\pgs, which relies on significantly more supervision, and handily outperforms any baseline implemented without \pg information.}%
    \vspace{-0.15in}

\end{figure*}

Our prior experiment demonstrated that \fname can solve a wide range of manipulation tasks. We now seek to understand if \fname can compete with baselines, which rely on significantly more expert supervision and tuning for stable exploration. Specifically, we consider three baselines (listed below) that broadly exemplify prior work in this area: 

% \VK{6/12: "Everything is a reward function" mindset is dominating in the section below}. Previous reviewers asked for more *specific* details in the text itself. Will try to balance here.
\begin{itemize}
    \item \textbf{DeepMimic}~\cite{peng2018deepmimic}: DeepMimic requires full hand and object trajectory supervision: it optimizes the robot to imitate expert fingertip poses in addition to the object trajectory. Thus, this baseline receives the maximum possible expert supervision at every time-step. DeepMimic is easily implemented by adding rewards and termination conditions for the fingertips to our existing task formulation. % To our knowledge this result is the \textit{first} time DeepMimic has been demonstrated on dexterous manipulation tasks. (may not be true anymore)
    \item \textbf{GRAFF$^*$}~\cite{mandikal2020dexterous}: GRAFF encourages the robot to make functional contacts with the objects using ``object affordances" -- i.e. parts of the object where a human expert would grasp to accomplish a task. In practice, it rewards the robot for making contact at ground truth grasping points. While the original paper operated on visual observations, we re-implement it with simulator state information for fair comparison. % \VK{27/1: Mention that their behaviors looks really bad. Add a few figures in appendix}
    \item \textbf{Task Curriculum}~\cite{karpathy2012curriculum}: This baseline uses an expert designed curriculum to accelerate policy learning, in the hope that learning easy tasks will accelerate learning harder tasks later on. First, the robot must learn how to stably pick (i.e. lift) objects. To learn the rest of the task, our full tracking objective (e.g. $\lambda_1$ from Sec.~\ref{sec:method_formulation}) is linearly activated over the course of ~4M timesteps (i.e. average time to learn lifting). 
\end{itemize}

\begin{figure*}[t]
    \centering
    \begin{subfigure}[b]{0.24\linewidth}
        \includegraphics[width=\linewidth]{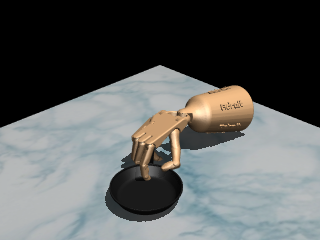}
        % \vspace{-0.12in}
        \caption{\small Pre-Grasp}
        \label{fig:ablate-default}
    \end{subfigure}
    \begin{subfigure}[b]{0.24\linewidth}
        \includegraphics[width=\linewidth]{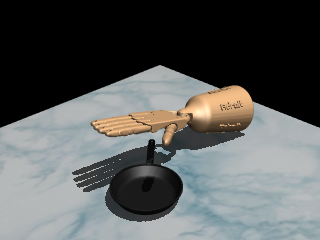}
        % \vspace{-0.12in}
        \caption{\small Open Hand}
        \label{fig:ablate-open}
    \end{subfigure}
    \begin{subfigure}[b]{0.24\linewidth}
        \includegraphics[width=\linewidth]{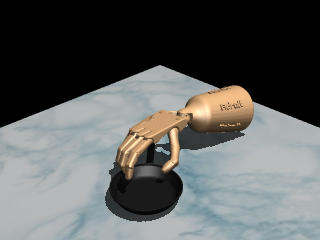}
        % \vspace{-0.12in}
        \caption{\small Mean Hand}
        \label{fig:ablate-mean}
    \end{subfigure}
    \begin{subfigure}[b]{0.24\linewidth}
        \includegraphics[width=\linewidth]{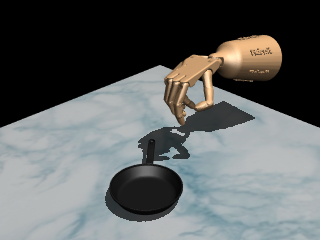}
        % \vspace{-0.12in}
        \caption{\small Distance Ablation}
        \label{fig:ablate-distance}
    \end{subfigure}
    \vspace{-0.05in}
    \caption{ Our ablations are implemented by sub-optimally adjusting the \pg pose. We show visualizations of this process with the ``Frying Pan" object.}
    \vspace{-0.05in}
    \label{fig:ablates}
\end{figure*}

\begin{table*}[t]
\centering
\resizebox{0.75 \linewidth}{!}{%
    \begin{tabular}{lccccc}
        \toprule
        % \rowcolor{lightgray}
         & \textbf{Pre-Grasps} & \multicolumn{2}{c}{\textbf{Ablate Pose}}  & \multicolumn{2}{c}{\textbf{Ablate Distance}}\\
         & & Open & Mean & 15 cm & 25 cm \\
         \midrule
         \textit{Success Metric} & \textbf{75.5\%} & $23.1\%$ & $59.5\%$  & $28.6\%$ & $19.7\%$\\
         
         \textit{Error Metric} & \textbf{4.82e-3} & $1.55$e-2 & $5.90$e-3 & $1.17$e-2 & $1.92$e-2\\
        \bottomrule
    \end{tabular}
}
\vspace{0.05in}
\caption{\fname is run on \suitename-30, using default and ablated \pgs. Error and success metrics at the end of training, are presented above. Destroying key \pg attributes (i.e. pose and proximity) significantly harms performance. The best ablation (Mean Hand), represents the least deviation from the default \pg.}

\label{tab:ablation}
\vspace{-0.2in}
\end{table*}

A major benefit of using \fname is that it shortens the task exploration horizon, since it positions the robot near the object. To control for this factor, we implement each baseline twice -- with and without \fname. Additional implementation details are presented in App.~\ref{appendix:baseline}. All six baselines (3 methods, w/ and w/out \fname) are evaluated against a \fname-only method on \suitename-30\footnote{\suitename-30 is used, since the baselines need added supervision.} tasks. Their performance at 50M steps are presented in Fig.~\ref{fig:baselinebar}. We observe that the baseline methods (which require dense supervision beyond \pgs) provide no appreciable performance boost (even when using \fname), and are completely ineffective w/out \fname. This is true even during the early stages of training: the \fname-only method remains competitive against all baseline w/ \fname implementations at every optimization step (see Fig.~\ref{fig:baselinecurve}).

These experiments offer strong evidence that \pgs act as crucial supervision for dexterous manipulation, since the baselines could only remain competitive when implemented w/ \fname. Simply put, behavior synthesis frameworks that leverage  \pgs can more easily acquire diverse dexterous manipulation behaviors, thus making further supervision far less valuable. Indeed, this observation is also reflected (though unacknowledged) in past learning work~\cite{chen2022system,huang2021generalization,akkaya2019solving} -- we find that removing \pgs from their setups causes them to collapse entirely (see App.~\ref{appendix:pg_related}).

\subsection{Ablations: What Makes a Pre-Grasp Useful?}
\label{sec:exp_ablations}

Our investigation has established that \pgs are a key source of supervision that enable scaling to the diverse tasks in \suitename. We now run an ablation study to understand what \pg properties (e.g. hand pose, proximity to object, etc.) make them useful during learning. %how sub-optimally modifying the \pg poses affects \fname's performance. That is to say, we seek to understand what properties of \pgs (e.g. hand pose, proximity to object, etc.) make them useful.

The following ablation classes are considered (visualized in Fig.~\ref{fig:ablates}):
\begin{itemize}
    
    \item \textbf{Ablate Pose:} This ablation tests if finger pose information (i.e. finger joint positions) is required for \fname to work. Specifically, we replace the \pg finger pose with both an ``Open Hand" (see Fig.~\ref{fig:ablate-open}) and a ``Mean Hand" (see Fig.~\ref{fig:ablate-mean}), calculated by averaging all the \pgs used in our investigation.
    \item \textbf{Ablate Distance:} The previous ablation does not address the importance of the object proximity. To test this factor, we shift the wrist away from the \pg (towards default robot's reset pose) by two fixed offsets (see Fig.~\ref{fig:ablate-distance}), while keeping the finger pose fixed.
\end{itemize}

These ablations\footnote{Also run on \suitename-30 for consistency w/ baselines.} (see Table.~\ref{tab:ablation}) reveal that object proximity matters significantly -- moving the hand away from the object dramatically reduces \fname's performance. Finger pose information is critical as well. The ``Open Hand" experiment demonstrates how removing the \pg's finger pose causes a drastic decrease in performance. Furthermore, the ``Mean Hand" experiments show that some of the fine-grained aspects of the \pg pose matter, since replacing it with a generic hand pose resulted in a $20\%$ performance hit. However, the Mean Hand does perform significantly better than the Open Hand setting ($59.5\%$ vs $23.1\%$ respectively), which indicates that \pgs need not be perfect for control. This error tolerance suggests that one could use predicted \pg states in policy learning.

\subsection{Real World Validation}

\begin{figure}
        \centering
        \includegraphics[width=0.9\linewidth]{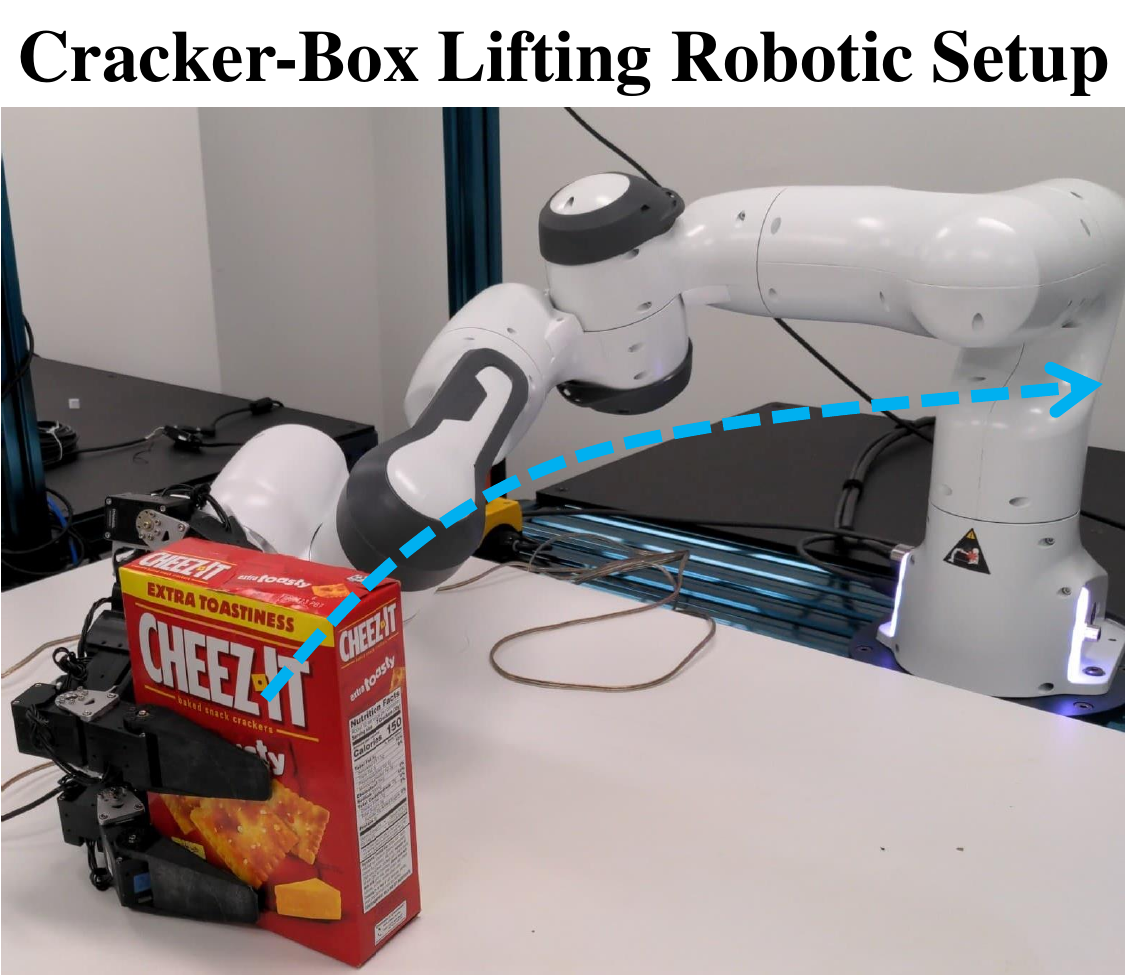}
        \caption{ Picture of our real world task setup. The D'Manus robotic hand is controlled to lift the ``Cheez-Itz" cracker-box from YCB~\cite{calli2015ycb}. }
        \label{fig:realworld}
        \vspace{-0.1in}
\end{figure}

Since \fname does not use extensive supervision to shape learning, it is possible that the learnt policies will not be viable on real hardware (e.g. actions are too aggressive). Our final experiment seeks to dispel this fear, by executing actions from a trained (using \fname) policy (in open-loop fashion) on an actual D'Manus robot. Specifically, a simple ``Cracker-box Lifting" task is defined using \fname, alongside a matching real-world environment replica (see Fig.~\ref{fig:realworld}, details in App.~\ref{appendix:realworld}). We find that simulated actions can be replayed on the robot -- i.e. the robot grasps and lifts the cracker-box using the learned behavior. This provides initial evidence that our simulated results could fully transfer to hardware. However, a more thorough real world investigation of our ideas (i.e. robot policy deployment for all tasks) is outside the scope of this paper. 

\section{Discussion} 
\label{sec:conclusion}
\vspace{-0.1in}

This paper demonstrated that simple ingredients can enable learning dexterous behaviors in diverse scenarios. Specifically, we use exemplar object trajectories as generic task specifiers, and \pgs as supervisory signals for exploration. Our primary contributions are (1) the \fname framework, which functions as a simple exploration prior for dexterous policy learning, and (2) the \suitename benchmark which fills an important gap -- the lack of diverse dexterous manipulation benchmarks (50 tasks, 20+ objects, 3 robots). Our system was able to achieve diverse control results with \textit{no} per-task expert engineering, while using the minimal possible supervision (i.e. a single \pg frame). Indeed, our learned behaviors match the performance of baselines that make significantly more assumptions. Finally, we characterize the \pg properties required for stable exploration, as well as demonstrate that our learned behaviors are physically plausible.

\section{Limitations and Future Work}
\label{sec:futurework}
\vspace{-0.1in}
% \VK{6/12: General Comment: Try to contrast to existing approaches a bit more and provide suggestions/ intuitions to look smart :) }

While our investigation was expansive, there are multiple vectors of improvement that should be addressed in future work. First, our investigation was primarily conducted in simulation, which is quite common in this space due to the lack of affordable dexterous hands (ShadowHand is \$100K+). However, a few affordable solutions are in development~\cite{gupta2021reset,ahn2020robel}, which we are starting to investigate. We hope to eventually deploy a fully trained \fname policy in the real world using a combination of: domain randomization~\cite{akkaya2019solving}; real world training~\cite{ahn2020robel,nagabandi2020deep}; and/or adaptation~\cite{miki2022learning}. In addition, the \pgs used in this investigation were curated, but this approach would not work ``in-the-wild" where objects are innumerable. Inspired by recent work in the vision community~\cite{ye2021shelf,cao2021reconstructing,liu2021semi,murali2020same}, we plan to replace curated \pgs with \pgs predicted from visual inputs using minimal assumptions (e.g. unknown object mesh). Next, while our trajectory centric task formulation can encode a wide range of behaviors, extensions are needed to further increase task diversity. For example, additional constraints will be needed for tasks with precise force requirements (e.g. hammering a nail with 15N force). In addition, we only consider single-object tasks without distractor objects or clutter. Handling these situations will require changes to our task formulation, and a more flexible (i.e. clutter-aware~\cite{laskey2016robot}) reaching policy in \fname. Finally, we hope to swap our policies from state to visual observations, in order to handle a wider range of (e.g. deformable) objects and enable inter-task generalization. 

\section*{Acknowledgments}
We'd like to acknowledge our collaborators at Carnegie Mellon and Meta AI Research in Pittsburgh, who gave valuable feedback that made the final paper much stronger. In particular, we'd like to recognize Shikhar Bahl, Homanga Bharadhwaj, Yufei Ye, and Sam Powers. Finally, this research was funded by Meta AI research and was conducted while the first author was an intern there.

%% Use plainnat to work nicely with natbib. 

\bibliographystyle{ieeetr}
\bibliography{references}

\clearpage
\appendix
The following Appendix sections are meant to add helpful context to our submission. In App.~\ref{appendix:pg}, we provide a more formal definition of \pgs (see App.~\ref{appendix:pg_intro}), contrast them against grasps (see App.~\ref{appendix:pg_vsgrasp}), and show how prior dexterous, learning algorithms relied on \pgs for stable performance (see App.~\ref{appendix:pg_related}). App.~\ref{appendix:hyper} presents all the hyper-parameters used in our task formulation (see App.~\ref{appendix:hyper_formulation}), \suitename benchmark (see App.~\ref{appendix:hyper_suite}), and \fname learning framework (see App.~\ref{appendix:hyper_framework}). App.~\ref{appendix:all_tasks} provides a more thorough breakdown of our experimental validation results (see Sec.~\ref{sec:exp_validate}). Finally, App.~\ref{appendix:baseline} gives more information about the baseline implementations, and App.~\ref{appendix:realworld} discusses our hardware setup, built using the Franka Panda and D'Manus~\cite{ahn2020robel} robots.

\subsection{Pre-Grasps in Depth}
\label{appendix:pg}

\begin{figure*}[!b]
    \centering
    \includegraphics[width=\linewidth]{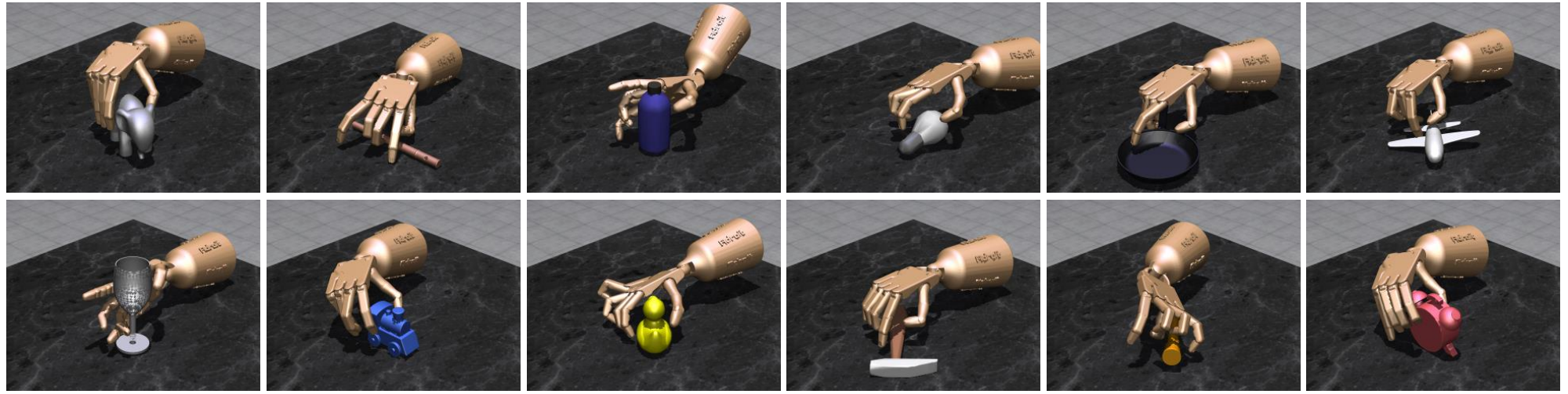} 
    \caption{Examples of \pgs used in our experiments. Note how they (1) place the hand in proximity to the target object and (2) and pose the fingers around functional parts of the object (e.g. hammer's handle). However, \pgs are \textit{not} stable grasps. In fact, the hand does not make contact with the objects at all!}
    
    \label{fig:pregrasp}
\end{figure*}

In this appendix we give a more formal introduction to \pgs (see App.~\ref{appendix:pg_intro}), and compare them to grasps (see App.~\ref{appendix:pg_vsgrasp}), which the reader may be more familiar with. Finally, we demonstrate that \pgs are often leveraged as an unstated assumption in prior work (see App.~\ref{appendix:pg_related}).

\subsubsection{What are Pre-Grasps?}
\label{appendix:pg_intro}

The pose of the hand (position, orientation, and joint articulation) just before the initiation of a hand-object interaction is called a ``\pg." It places the hand in a favourable region of the state space relative to the object, so that the intermittent contact behaviors of dexterous manipulation can evolve stably. Note that we did not develop this concept: it is a classical robotics construct~\cite{kang1994determination,kappler2010representation,kappler2012templates}.

We now outline a few key properties of \pg states (shown in Fig.~\ref{fig:pregrasp}) -- (1) First, a \pg positions the robot close to the target object, and orients the robot's palm and wrist joints towards the object. This proximity ensures that \pgs can easily evolve into a stable grasp, without requiring the robot to explore the whole state space. (2) In addition, \pg finger poses encode valuable information about functional parts of an object, without requiring the robot to reason about it explicitly. For example, a \pg that curls a robot's fingers around a mug handle, offers a crucial signal for the robot to interact with the mug by grasping the handle. This property also implies that there might be multiple \pgs possible for every object (corresponding to different functions). (3) Finally, \pg states incentivize  favorable contacts (e.g. interaction with tool handles) and avoid dangerous contacts with the object (e.g. knife edge) and/or any other parts of the scene (e.g. pressing into the table). This is crucial because dexterous manipulation is full of contacts that are difficult to effectively model, predict, and reason about. A good \pg provides a favourable start and strong momentum for learning the downstream manipulation behavior.

\subsubsection{Grasp vs Pre-Grasp}
\label{appendix:pg_vsgrasp}

Traditionally, successfully grasping an object is considered to be the most important event for dexterous manipulation. Indeed, dexterous manipulation has often been defined as a series of grasps in sequential order~\cite{miller2004graspit,howard1996stability,bicchi1995closure}. The first grasp in the sequence has been extensively analyzed both for grasp synthesis \cite{miller2004graspit} and for object stability \cite{howard1996stability, bicchi1995closure}. Therefore, it begs an important question: are \pgs or grasps the key states for dexterous manipulation? We believe that \pgs are a stronger and more practical primitive to leverage when compared to grasps. They are stronger because if \pgs can simplify policy learning, then grasps should as well. However, the reverse is not true, since grasping comes after \pgs. Furthermore, they are more practical, since reaching a stable grasp is significantly harder than reaching a \pg. After all, grasps are based on precise geometric and surface properties (friction, softness, etc) that cannot be accurately detected with modern position and force sensing technologies. Furthermore, grasps have extremely narrow stability margins that can be violated by small deviations. 

Our key insight is that a grasp is a ``post-condition" (effect) of gaining control of the object. However, to make dexterous manipulation policy learning feasible, we need a ``pre-condition" (which is easier to reach than ``post-condition") that can initialize the manipulation behavior in a healthy manner. We hypothesize that \pgs (i.e. hand pose before the onset of object interaction) are the key pre-condition, and arriving/initializing at a \pg can significantly lower the complexity in synthesizing dexterous manipulation behaviors. Unlike grasps, \pgs -- (1) are devoid of precise hand-object contacts and therefore doesn't have extreme sensing requirements; (2) are easy to achieve and satisfy as most robots have good joint position control; (3) have a wide stability basin such that various contact transitions needed for dexterous manipulation can evolve within its boundaries.

\subsubsection{Pre-Grasps in Past Learning Work}
\label{appendix:pg_related}

We note that many dexterous manipulation papers in the last decade used \pgs to some degree without explicit mention. For example, the Pen-Task in DAPG's ADROIT suite~\cite{rajeswaran2017learning} required \pg initialization to work. Similarly, the OpenAI Rubik's cube experiment~\cite{akkaya2019solving} assumed a stable in-hand (i.e. \pg) initialization.

Two recent dexterous manipulation papers~\cite{chen2022system,huang2021generalization} considered the in-hand object rotation task. For most of their experiments they start at a \pg pose (object laid flat on the palm). As a test, we experimentally demonstrate for Huang et. al~\cite{huang2021generalization} this choice is crucial -- removing the \pg initialization causes learning to fail entirely. For Chen et. al.~\cite{chen2022system} this choice is explicitly addressed in the paper -- the only way they were able to avoid starting at \pg was by using a \textit{gravity curriculum}. Such a procedure would be impossible in the real world, where gravity is a fixed constant.

\begin{table*}[!t]%\[!htb\]
    \begin{subtable}[t]{.37\textwidth}
        \centering
            \begin{tabular}{@{\extracolsep{1pt}}lc}
                \hline \\
                \multicolumn{2}{c}{\textbf{Task Formulation  Parameters}}\\
                \hline \\
                Parameter & Value \\
                \hline \\
                $\lambda_1$ & 10 \\
                $\lambda_2$ & 2.5 \\
                $\alpha$ & 50 \\
                $\beta$ & 5  \\
                $\zeta$ & 0.02  \\
                $\omega$  & 0.25  \\
                -- & -- \\
                -- & -- \\
                -- & -- \\
                -- & -- \\
                -- & -- \\
                -- & -- \\
                \hline \\
            \end{tabular}
    \end{subtable}
   \begin{subtable}[t]{.63\textwidth}
        \centering
            \begin{tabular}{@{\extracolsep{1pt}}lc}
                \hline \\
                \multicolumn{2}{c}{\textbf{\fname Framework Parameters}}\\
                \hline \\
                Parameter & Value \\
                \hline \\
                $\gamma$ (discount) & 0.95 \\
                $\lambda$ (GAE) & 0.95 \\
                Learning Rate & 1e-5  \\
                Value Fn. Coef. & 0.5  \\
                Entropy Coef. & 0.001  \\
                PPO Clip & 0.2  \\
                Steps per Iter & 4096  \\
                Mini-Batch Size & 256  \\
                Epochs & 5 \\
                Policy Net ($Mean$) & MLP([256, 128]); TanH + Ortho. Init.\\
                Policy Net ($\sigma$) & Param(size=\textit{adim}, value=exp(-1.6)) \\
                Value Fn. Net & MLP([256, 128]);  TanH + Ortho. Init.\\
                \hline \\
            \end{tabular}
    \end{subtable}
    \caption{Experimental hyper-parameters for our task creation and policy learning stacks. Note that no per-task tuning was allowed at all for any of the listed parameters! }
    \label{tab:hparams}
\end{table*}
\subsection{Hyper-Parameters}
\label{appendix:hyper}

In this appendix we list the hyper-parameters for our task formulation (see App.~\ref{appendix:hyper_formulation}), \suitename benchmark (see App.~\ref{appendix:hyper_suite}), simulated environments (see App.~\ref{appendix:hyper_sim}), and \fname learning framework (see App.~\ref{appendix:hyper_formulation}).

\subsubsection{Task Formulation Hyper-Parameters}
\label{appendix:hyper_formulation}

As discussed in Sec.~\ref{sec:method_formulation}, our task formulation automatically parameterizes task MDPs using exemplar object trajectories. This is accomplished (in part) by using a simple reward function and termination function. The hyper-parameters for these functions are listed in Table~\ref{tab:hparams} (left).

\subsubsection{\suitename Tasks}
\label{appendix:hyper_suite}

As discussed in Sec.~\ref{sec:implement_suite}, we leverage our task formulation to automatically generate a benchmark -- \suitename. Recall that each task in \suitename is parameterized using an exemplar object trajectory (more info in Sec.~\ref{sec:method_formulation} and App.~\ref{appendix:hyper_formulation}). The trajectories for these tasks were mined from three sources listed below:
\begin{itemize}
    \item \textbf{Human Motion Capture (GRAB~\cite{taheri2020grab}):} The GRAB data-set contains motion capture sequences of human-object interactions. Specifically, each recording in GRAB consists of a human performing various skills with a collection of common objects (e.g. use hammer, pass flute, lift duck, etc.) from the ContactDB dataset~\cite{brahmbhatt2019contactdb}. The recordings contain the target object's pose (i.e. position and orientation), and a mesh reconstruction of the articulated hand at each time-step. In other words, GRAB contains paired exemplar object trajectories ($X$ from Sec.~\ref{sec:method_formulation}) and hand pose trajectories $H = [h_0, \dots, h_n]$. Note that these hand trajectories are not used in our method, but are required for the baselines. We created 40 tasks using trajectories from GRAB.
    \item \textbf{Expert Policies (ADROIT~\cite{rajeswaran2017learning}):} Another source of object trajectories are expert policies, acquired either by learning/writing a controller or through human expert tele-operation. Since expert policies produce object behaviors worth imitating, one can simply extract the object trajectory from successful policy roll-outs. Specifically, we take expert trajectories from the ADROIT benchmark suite, which were collected by tele-operating a simulated ShadowHand. ADROIT object trajectories (no actions!) were taken to parameterize 3 tasks. 
    \item \textbf{Human Animators:} Finally, we showcase that object trajectories for our method need not come from data. Instead, suitable (i.e. smooth) object trajectories can be scripted by human animators. We manually animated 7 object trajectories and used them to create the final tasks.
\end{itemize}

In addition \pgs for each task are mined from the following sources:
\begin{itemize}
    \item \textbf{MoCap:} We extract human hand poses from the GRAB MoCap dataset (discussed above) and transfer them to the robot using an inverse kinematics procedure (solve for joints that achieve human finger-tip positions). This allows us to easily create \pgs for trajectories sourced from MoCap datasets.
    \item \textbf{Tele-Op:} Expert Tele-Op trajectories provide a natural source for \pgs. We extract \pg states from the dataset provided by the DAPG paper~\cite{rajeswaran2017learning}.
    \item \textbf{Human Labels:} These \pgs are manually labeled by a human annotator. This is accomplished using the MuJoCo visualizer UI. 
    \item \textbf{Learned Model:} We feed the object mesh and pose into the GrabNet model~\cite{taheri2020grab}, which in turn predicts a human grasp pose (w/ MANO parameters). This pose is transferred to the robot using inverse kinematics in the same fashion as the MoCap \pgs.
\end{itemize}

Altogether, we created 50 unique tasks using these trajectory sources, and bundled them together into the \suitename benchmark. However, the the baselines required extra supervision (i.e. full hand reconstruction) that was only present in the GRAB data-set. In addition, 10 of the tasks constructed using GRAB data focused on ``simple" lifting behaviors that (while good for debugging) were not representative of free-form object motion. Thus, we created the \suitename-30 benchmark for our baseline study that (as the name suggests) consisted of the remaining 30 \suitename tasks parameterized w/ GRAB data. Table~\ref{tab:tasks} lists all of the tasks, alongside the object trajectory source, \pg sources, and \fname performance. Please refer to our code release for benchmark membership (i.e. in \suitename-30 or not) for each task.

\subsubsection{Implementing \suitename w/ a Physics Simulator}
\label{appendix:hyper_sim}
Prior sections (Sec.~\ref{sec:method_formulation},~\ref{sec:implement_suite} and App.~\ref{appendix:hyper_formulation},~\ref{appendix:hyper_framework}) have discussed how \suitename tasks are formulated, the hyper-parameters used, the trajectories that parameterize them, and the design decisions made while building them. In contrast, this section will describe how \suitename tasks are actually implemented.

Our investigation leveraged the MuJoCo~\cite{todorov2012mujoco} physics simulator in place of a real robotics setup. Simulation was used since dexterous hardware is hard to acquire, and because real setups are less reproducible. For each task, we built a simulated scene that contains a table, a single robot, and a single target object. Object matching was done to explicitly pair exemplar trajectories with the intended object (e.g. drinking trajectory is matched w/ mug), while the robot matching was done arbitrarily. We consider 3 robot morphologies -- ShadowHand~\cite{Kumar2016thesis}, D'Hand~\cite{gupta2021reset}, and D'Manus~\cite{ahn2020robel} -- alongside 34 objects from the ContactDB~\cite{brahmbhatt2019contactdb} and YCB~\cite{calli2015ycb} object sets. Note that the simulated object properties are carefully defined to avoid inconsistencies. For the ContactDB objects, we infer object properties (e.g. mass and moments of inertia) by adopting common properties from 3D printed objects (ContactDB objects are printed). This was achieved by setting the object density to $1.25$ g/cm$^3$ (density of PLA) and then using MuJoCo to solve necessary object properties from the convex decomposition of the objects. The convex decomposition was solved using VHACD~\cite{VHACD}. The YCB object properties are defined by measuring real world versions of the objects.

We've now described the simulated setup and MDP formulation for every \suitename task. These are implemented together into a single Gym environment~\cite{gym}. As a result, our tasks can be easily ``plugged" into other RL and control code-bases. We now describe the observation space and action space used in our environment:
\begin{itemize}
    \item \textbf{Observation Space:} We use a simple state space consisting of the robot joints, robot fingertip locations, and the object pose. In addition, we use positional encoding~\cite{vaswani2017attention} to mark the current simulation time-step and add that to the end of the state vector. This is needed to handle time varying goal behaviors (see Sec.~\ref{sec:method_formulation}).
    \item \textbf{Action Space:} Our action space is joint position control, achieved using a simple low level PD torque controller. The first six joints handle wrist position/orientation, and the remaining joints control the robot fingers themselves. The simulated gains are tuned to be realistic, and gravity compensation is applied (by modifying applied torques). These assumptions are all consistent with real dexterous hand hardware.
\end{itemize}
Please check our website for the code and further documentation of \suitename tasks and environments: \website.

\subsubsection{Learning Behaviors w/ \fname}
\label{appendix:hyper_framework}

As discussed in Sec.~\ref{sec:implement_fname}, our \fname framework uses \pg states as exploration primitives to speed up dexterous behavior learning. It operates in two phases: (1) the robot hand is brought to a \pg state using a heuristic trajectory optimization planner; and (2) the behavior policy is learned by optimizing task reward (e.g. from \suitename) using a RL algorithm. Both ingredients are described in detail below. Furthermore, \fname hyper-parameters are listed in Table.~\ref{tab:hparams} (right), and psuedo-code is shown in Alg.~\ref{alg:pgdm}.

\paragraph{Reaching Pre-Grasp States:} After the task scene is reset (e.g. object brought to reset position), \fname's first stage begins. Specifically, the robot hand is brought to an appropriate \pg position for the task and target object. Note that the \pg state is not predicted, instead it is manually annotated (per-task) in an offline dataset. A CEM trajectory planner~\cite{Lowrey-ICLR-19} solves for actions that bring the robot to the \pg, without disturbing the object. This is done in a geometry-free way (i.e. w/out any object mesh knowledge). First, the hand is brought far above (e.g. 30cm) the target object. Second, the thumb joint begins moving towards its \pg configuration. And finally, the rest of the hand is brought to the \pg state. While a simple heuristic, this motion plan successfully brings the robot to \pgs for \textit{all} of the 50 diverse tasks (and 34 objects) considered in \suitename, without \textit{any} scene/geometry information. This will not scale to more complicated settings (e.g. reaching \pg states in clutter), but those situations are beyond the scope of this paper.

\paragraph{Behavior Learning w/ Pre-Grasps:} Once at a \pg state, \fname swaps to a policy learning algorithm, to optimize the original task reward. Specifically, we utilize the PPO~\cite{schulman2014motion} policy-gradient RL algorithm for policy learning. Our code uses the Stable-Baselines3~\cite{stable-baselines3} implementation that is built upon the Pytorch~\cite{paszke2019pytorch} deep learning library. Please check Table.~\ref{tab:hparams} for the exact RL hyper-parameters. 

\begin{algorithm}[!t]
\caption{Learning \suitename tasks w/ \fname}
\label{alg:pgdm}
\algcomment{\fontsize{7.2pt}{0em}\selectfont \texttt{l2error}: euclidean distance;  \texttt{absang}: mag. of quat. angle; \texttt{exp}: natural exponent.
%\vspace{-1.em}
}
\definecolor{codeblue}{rgb}{0.25,0.5,0.5}
\lstset{
  backgroundcolor=\color{white},
  basicstyle=\fontsize{7.2pt}{7.2pt}\ttfamily\selectfont,
  columns=fullflexible,
  breaklines=true,
  captionpos=b,
  commentstyle=\fontsize{7.2pt}{7.2pt}\color{codeblue},
  keywordstyle=\fontsize{7.2pt}{7.2pt},
%  frame=tb,
}
\begin{lstlisting}[language=python]
####################### INPUTS ###########################
# policy: Randomly initialized policy network ($\pi$)
# CEM: Trajectory Planner for reset policy
# PPO: Implementation of PPO algorithm (e.g. SB3)
# goals, pre_grasp: exemplar trajectory and pregrasp state
###################### CONSTANTS #########################
# N_ITERS, N_STEPS: PPO iterations/rollout buf size
# L_1, L_2, A, B, Z: constants for Eqn. (1)
# gamma: termination bound for Eqn. (2)
##########################################################

class TCDMEnv(gym.Env)
    .....
    
    # reward helper function
    def reward_fn(self)
        state = self.get_state()
        t = state.time
        g = self.goal[t]
        
        # calculate error terms
        delta_xyz = l2error(state.obj_xyz - g.obj_xyz)
        delta_ori = absang(state.obj_ori, g.obj_ori)
                
        # calculate reward with Eqn. (1)
        R = L_1 * exp(-A * delta_xyz - B * delta_ori)
        if next_s.obj_z > Z and g.obj_z > Z:
            R += L_2
        return R
    
    # termination helper function
    def termination_fn(self):
        state = self.get_state()
        t = state.time

        # terminate if error exceeds threshold
        delta_xyz = l2error(state.obj_xyz - g.obj_xyz)
        return delta_xyz > gamma


# helper function for rolling out policy 
def get_rollouts(policy, n_steps):
    # aggregates data from the rollouts
    buffer = RolloutBuffer() 
    
    while len(buffer) < n_steps:
        # reset environment and move robot hand to pre_grasp
        s = env.reset()
        reset_policy.moveto(env, pre_grasp)
        
        while True:
            # get action from policy and step environment
            a = policy.sample(s, g)
            next_s, r, done, info = env.step(a)
            
            # append to buffer and advance state
            buffer.append(s, a, next_s, R)
            s = next_s
            
            # reset episode if done
            if done:
                break
    return buffer


# load task and reset planner
env = TCDMEnv(*)
reset_policy = CEM()

# train policy with simple PPO update loop
env = TCDMEnv(*)
for i in range(N):
    eval_and_log(policy)
    buffer = get_rollouts(policy, N_STEPS)
    PPO.update(policy, buffer)
\end{lstlisting}

\end{algorithm}

% policy learning framework hyperparameters

\subsection{ \fname Validation Experiment Breakdown }
\label{appendix:all_tasks}

\begin{figure*}
    \centering
    \includegraphics[width=\linewidth]{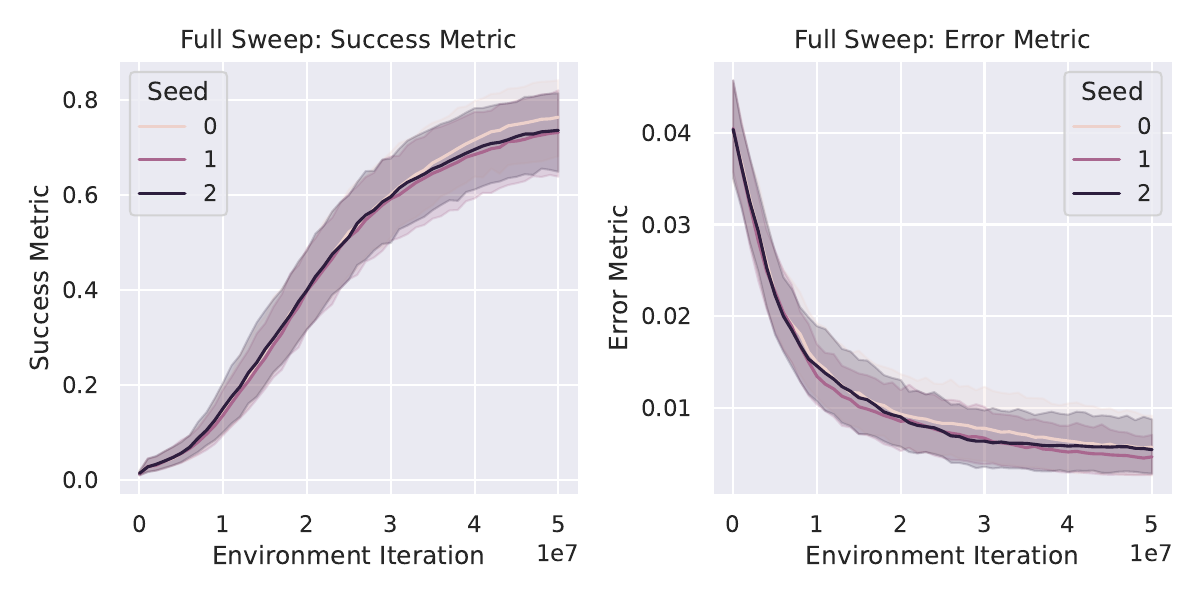}

    \caption{ These learning curves show how \fname's error and success metrics (averaged across all \suitename tasks) evolve over RL training. Our experiments were run with three seeds, each plotted separately. Note how \fname effectively learns the tasks with low run to run variance. }
    \vspace{-0.2in}
    \label{fig:scalingcurves}
\end{figure*}

\begin{figure*}
    \centering
    \includegraphics[width=\linewidth]{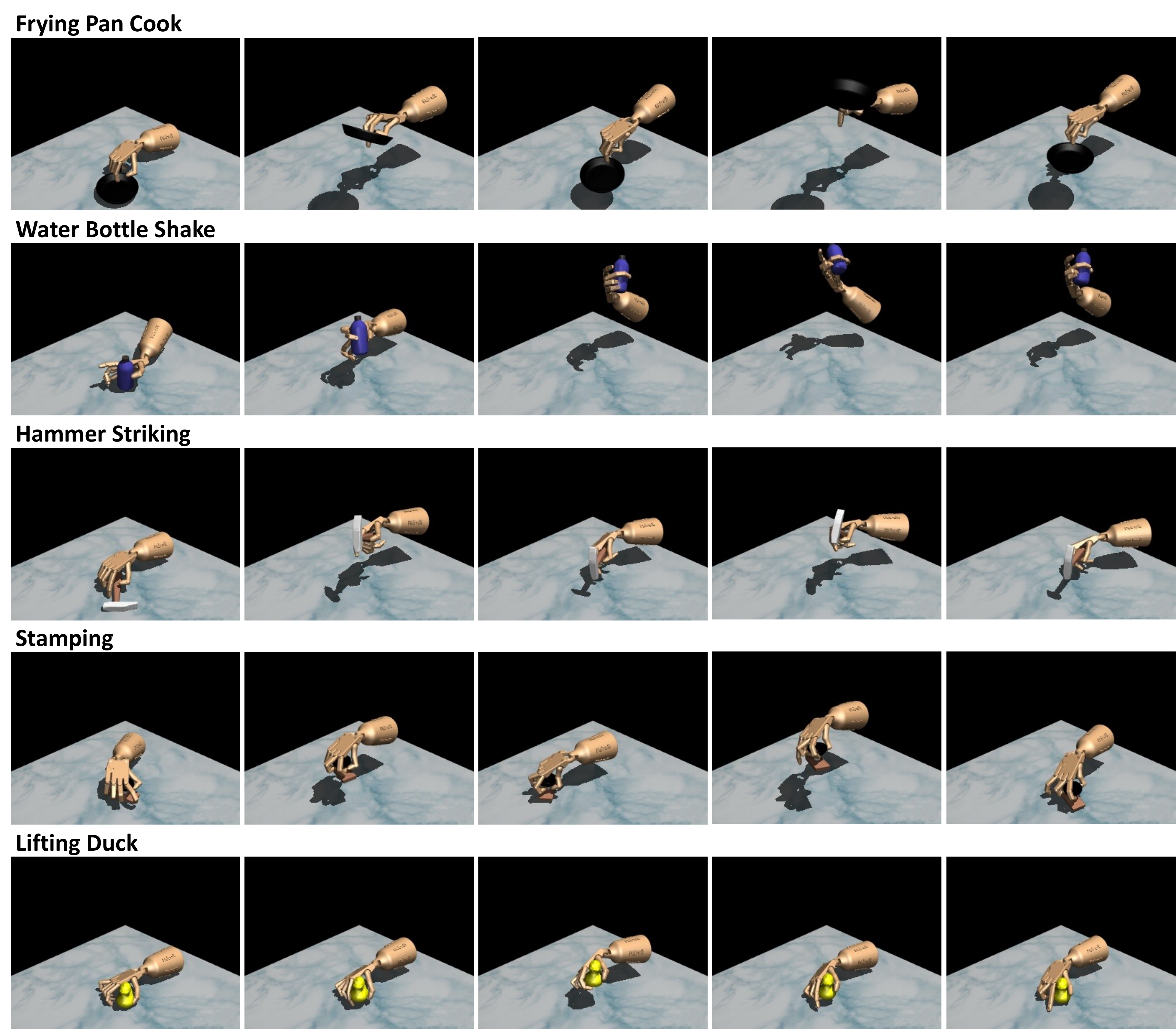} 
    \caption{Qualitative results of our learned policies on select tasks from \suitename. Note how \suitename contains diverse objects with behaviors ranging from shaking, to hammering, to lifting. Furthermore, our policies learn natural ``human like" actions, despite never optimizing for them. For animated visualizations please refer to our website: \website. }%
    \vspace{-0.2in}
    \label{fig:qualitative}
\end{figure*}

This appendix presents additional context and results from the full validation experiment that could not fit in the main paper. Recall (from Sec.~\ref{sec:exp_validate}) that we deployed our \fname framework on all 50 tasks in \suitename using 3 random seeds per task. Performance by task is broken down in Table~\ref{tab:tasks}, and learning curves are presented in Fig.~\ref{fig:scalingcurves}. Additionally, qualitative examples of the behaviors are shown in Fig.~\ref{fig:qualitative}. Note the diversity of our task suite, and how \fname works across a wide variety of scenarios, with no fine-tuning and minimal variance between seeds. For animated visualizations, please check our website: \website.

\subsection{Baseline Implementations}
\label{appendix:baseline}

This appendix provides added context for the DeepMimic (see App.~\ref{appendix:baseline_deep}), GRAFF$^*$ (see App.~\ref{appendix:baseline_graff}), and Task Curriculum (see App.~\ref{appendix:baseline_curr}) methods used as baselines in our experiments (see Sec.~\ref{sec:exp_baselines}). Note that these baselines require added hand supervision that is only found in the GRAB data-set (see App.~\ref{appendix:hyper_suite}). As a result, our baselines are run on only a subset of tasks (i.e. \suitename-30).

\subsubsection{DeepMimic}
\label{appendix:baseline_deep}

DeepMimic~\cite{peng2018deepmimic} uses additional rewards to supervise the hand motion in addition to the object motion. Specifically, given an expert hand trajectory -- $H = [h_0, \dots, h_T]$ where $h_i$ is a hand-pose (i.e. joint angles) at step $t$ -- and a matching exemplar object trajectory $X$ (see Sec.~\ref{sec:method_formulation}), the robot is trained to match the expert trajectory and exemplar object trajectory at the same time. This is done with added (in addition to \suitename-30) reward and termination condition, $\lambda_3 exp(-\eta || h_t - \hat{h}_t ||_2)$ and $|| h_t - \hat{h}_t ||_2 > \gamma_h$ respectively, where $\hat{h}_t$ is the achieved robot pose at step $t$. We set $\lambda_3=1,\eta=15,\gamma_h=0.1$, since that gave best performance. Note that the original DeepMimic implementation used additional reward terms, which we dropped since they didn't change performance in our setting. Additionally, $H$ is calculated by applying inverse kinematics to the original hand pose trajectories from GRAB.

\subsubsection{GRAFF$^*$}
\label{appendix:baseline_graff}

The original GRAFF~\cite{mandikal2020dexterous} baselines predicted contact affordance regions on objects and encouraged the robot to interact with them. However, this original implementation used visual observations, whereas our setup uses robot state information. To make comparison fair, we extracted ground truth contact locations on the object from the GRAB dataset. In other words, we found where the human expert made contact with the object during MoCap recording, and thus created an optimal contact pattern for the robot to match. We then added a reward that incentivizes the robot to move its fingertips to those contact locations -- $\lambda exp(-\eta || c_t - \hat{f}_t ||_2)$ where $c_t$ is desired contact at step $t$ and $\hat{f}_t$ is robot fingertips. We found $\lambda=1,\eta=15$ gave best performance. This reward was added on top of the standard (e.g. \suitename-30) reward. To summarize, we re-implemented GRAFF using state-only observations, and made the comparison fair by adding expert (e.g. optimal fingertip location) data. 

\subsubsection{Task Curriculum}
\label{appendix:baseline_curr}

The task curriculum is our simplest baseline. We start by training the robot to only lift the object (i.e. satisfy the lifting bonus from Sec.~\ref{sec:method_formulation}). The hope is that the robot will be able to easily learn the desired behavior (e.g. toast wineglass) once it knows how to lift the target object. Thus, the object matching losses are activated over the course of training. This can be done by linearly blending $\lambda_1$ (see Sec.~\ref{sec:method_formulation}) from 0 to 1 over 4 million environment steps. This induces a curriculum into the \suitename tasks.

\subsection{Real World}
\label{appendix:realworld}

This appendix describes our real world verification experiment (see Sec.~\ref{sec:exp_validate}) in further depth. The goal here is to demonstrate that actions predicted by \fname policies can be deployed on real hardware. We do \textbf{not} consider training \fname policies in the real world, nor do we fully transfer closed-loop control policies from sim to the real world.  

\paragraph{Task Setup} Our task is to use a D'Manus robot, which is a low-cost hand from RoBEL suite~\cite{ahn2020robel}, to lift the Cheez-Itz cracker-box object from the YCB object suite~\cite{calli2015ycb}. A suitable task automatically defined using our MDP formulation (see Sec.~\ref{sec:method_formulation}). Specifically, we use a human animated ``object lifting" exemplar trajectory, which lifts and holds the box $0.2$m off the table, to define the task. In addition, we annotate an appropriate \pg state, with the object positioned between the D'Manus outstretched fingers and thumb (see Fig.~\ref{fig:realworld}), for the task.

\paragraph{Simulated Policy Learning} After defining the task, we created a simulated environment for it like normal (see App.~\ref{appendix:hyper_sim}), and learn a behavior policy to solve it using \fname. The learning procedure is almost exactly the same as our other simulated experiments, except for the addition of \textit{domain randomization}. Specifically, at every environment reset the cracker-box's width, mass, and friction are randomized within a pre-specified range. This is done to make the policy robust across a range of realistic possibilities, since the physical cracker-box will not perfectly match its simulated counterpart.

\paragraph{Real World Playback} After training the policy, we extract roll-outs from it in simulation and execute the actions taken in sim on a matching real world setup. Specifically, our real world setup uses a tabletop scene, with a YCB cracker-box target object, and a D'Manus robot mounted on a Franka Panda arm (see Fig.~\ref{fig:realworld}). For the D'Manus hand action playback happens in a straightforward manner: the learned actions are are simply replayed on the real world hardware using a PD controller. However, the base joints are modelled in sim using a free-joint with attached $x,y,z,\theta_x,\theta_y,\theta_z$ degrees of freedom. While this is suitable for sim, it is not achievable in the real world. To overcome this issue, we extract a base joint trajectory (e.g. the commanded base positions over time), convert it into Franka joint space using inverse kinematics~\cite{tunyasuvunakool2020}, and play those joints back on the real Franka robot using a PD controller from Polymetis~\cite{Polymetis2021}. The Franka actions are played in real-time alongside the D'Manus actions, resulting in smooth action playback for both the base and fingers.

\paragraph{Results} As discussed in Sec.~\ref{sec:exp_validate}, we are able to successfully lift the cracker-box object using the actions optimized in simulation. No real world adaptation or fine-tuning was required to make this possible (videos on our website\footnote{\website}). This suggests that our simulated results are plausible in the real world. In addition, we demonstrate that the learned actions are not too aggressive for robot hardware, despite the lack of any additional human supervision. However, this result does not prove that our actual policies can be transferred onto hardware. Achieving real world transfer will require significant engineering outside the scope of this paper (see Sec.~\ref{sec:futurework}). 

\begin{table*}[!t]%\[!htb\]
    \centering
    \begin{tabular}{lcccc}
        \toprule
        Task & Pre-Grasp Source & Trajectory Source & Success & Error (m) \\
        \midrule
        \textbf{airplane-fly1} & MoCap & MoCap~\cite{taheri2020grab} & 71.7\% & 6.66e-03 \\
        \textbf{airplane-pass1} & MoCap & MoCap~\cite{taheri2020grab} & 60.7\% & 2.69e-03 \\
        \textbf{alarmclock-lift} & MoCap & MoCap~\cite{taheri2020grab} & 41.8\% & 7.39e-03 \\
        \textbf{alarmclock-see1} & Learned & MoCap~\cite{taheri2020grab} & 85.7\% & 1.39e-03 \\
        \textbf{banana-pass1} & MoCap & MoCap~\cite{taheri2020grab} & 99.7\% & 4.11e-04 \\
        \textbf{binoculars-pass1} & MoCap & MoCap~\cite{taheri2020grab} & 74.3\% & 2.35e-03 \\
        \textbf{cup-drink1} & MoCap & MoCap~\cite{taheri2020grab} & 97.6\% & 3.66e-04 \\
        \textbf{cup-pour1} & MoCap & MoCap~\cite{taheri2020grab} & 83.1\% & 8.80e-04 \\
        \textbf{dhand-alarmclock} & Labeled & Animator & 84.8\% & 8.85e-04 \\
        \textbf{dhand-binoculars} & Labeled & Animator & 13.7\% & 3.80e-02 \\
        \textbf{dhand-cup} & Labeled & Animator & 85.5\% & 3.13e-03 \\
        \textbf{dhand-elephant} & Labeled & Animator & 10.0\% & 4.19e-02 \\
        \textbf{dhand-waterbottle} & Labeled & Animator & 70.1\% & 2.48e-03 \\
        \textbf{dmanus-coffeecan} & Labeled & Animator & 62.9\% & 1.46e-03 \\
        \textbf{dmanus-crackerbox} & Labeled & Animator & 98.0\% & 6.28e-04 \\
        \textbf{door-open} & Tele-Op & Expert~\cite{rajeswaran2017learning} & 92.2\% & 5.02e-04 \\
        \textbf{duck-inspect1} & Learned & MoCap~\cite{taheri2020grab} & 99.3\% & 3.94e-04 \\
        \textbf{duck-lift} & MoCap & MoCap~\cite{taheri2020grab} & 97.2\% & 3.00e-04 \\
        \textbf{elephant-pass1} & MoCap & MoCap~\cite{taheri2020grab} & 50.5\% & 1.58e-02 \\
        \textbf{eyeglasses-pass1} & MoCap & MoCap~\cite{taheri2020grab} & 39.4\% & 1.73e-02 \\
        \textbf{flashlight-lift} & MoCap & MoCap~\cite{taheri2020grab} & 97.1\% & 6.53e-04 \\
        \textbf{flashlight-on2} & MoCap & MoCap~\cite{taheri2020grab} & 94.9\% & 5.13e-04 \\
        \textbf{flute-pass1} & MoCap & MoCap~\cite{taheri2020grab} & 65.3\% & 7.71e-03 \\
        \textbf{fryingpan-cook2} & MoCap & MoCap~\cite{taheri2020grab} & 98.7\% & 3.75e-04 \\
        \textbf{hammer-strike} & Tele-Op & Expert~\cite{rajeswaran2017learning} & 64.0\% & 2.50e-03 \\
        \textbf{hammer-use1} & MoCap & MoCap~\cite{taheri2020grab} & 99.3\% & 4.11e-04 \\
        \textbf{hand-inspect1} & MoCap & MoCap~\cite{taheri2020grab} & 97.1\% & 1.02e-03 \\
        \textbf{headphones-pass1} & MoCap & MoCap~\cite{taheri2020grab} & 55.3\% & 1.14e-02 \\
        \textbf{knife-chop1} & MoCap & MoCap~\cite{taheri2020grab} & 93.4\% & 7.30e-04 \\
        \textbf{lightbulb-pass1} & MoCap & MoCap~\cite{taheri2020grab} & 43.7\% & 3.13e-02 \\
        \textbf{mouse-lift} & MoCap & MoCap~\cite{taheri2020grab} & 57.2\% & 6.39e-03 \\
        \textbf{mouse-use1} & MoCap & MoCap~\cite{taheri2020grab} & 21.7\% & 2.82e-03 \\
        \textbf{mug-drink3} & MoCap & MoCap~\cite{taheri2020grab} & 35.4\% & 1.39e-02 \\
        \textbf{piggybank-use1} & MoCap & MoCap~\cite{taheri2020grab} & 1.6\% & 6.06e-03 \\
        \textbf{scissors-use1} & MoCap & MoCap~\cite{taheri2020grab} & 87.4\% & 8.12e-04 \\
        \textbf{spheremedium-lift} & MoCap & MoCap~\cite{taheri2020grab} & 100.0\% & 1.61e-04 \\
        \textbf{spheremedium-relocate} & Tele-Op & Expert~\cite{rajeswaran2017learning} & 97.3\% & 3.68e-04 \\
        \textbf{stamp-stamp1} & MoCap & MoCap~\cite{taheri2020grab} & 44.9\% & 1.18e-02 \\
        \textbf{stanfordbunny-inspect1} & Learned & MoCap~\cite{taheri2020grab} & 86.2\% & 2.88e-03 \\
        \textbf{stapler-lift} & MoCap & MoCap~\cite{taheri2020grab} & 60.8\% & 5.16e-03 \\
        \textbf{toothbrush-lift} & MoCap & MoCap~\cite{taheri2020grab} & 100.0\% & 3.03e-04 \\
        \textbf{toothpaste-lift} & MoCap & MoCap~\cite{taheri2020grab} & 100.0\% & 2.80e-04 \\
        \textbf{toruslarge-inspect1} & MoCap & MoCap~\cite{taheri2020grab} & 70.7\% & 5.15e-03 \\
        \textbf{train-play1} & MoCap & MoCap~\cite{taheri2020grab} & 99.5\% & 2.78e-04 \\
        \textbf{watch-lift} & MoCap & MoCap~\cite{taheri2020grab} & 97.8\% & 5.99e-04 \\
        \textbf{waterbottle-lift} & MoCap & MoCap~\cite{taheri2020grab} & 51.2\% & 1.74e-03 \\
        \textbf{waterbottle-shake1} & MoCap & MoCap~\cite{taheri2020grab} & 92.4\% & 1.03e-03 \\
        \textbf{wineglass-drink1} & MoCap & MoCap~\cite{taheri2020grab} & 99.8\% & 1.82e-04 \\
        \textbf{wineglass-drink2} & MoCap & MoCap~\cite{taheri2020grab} & 100.0\% & 1.94e-04 \\
        \textbf{wineglass-toast1} & MoCap & MoCap~\cite{taheri2020grab} & 92.1\% & 1.04e-03 \\
        \bottomrule
    \end{tabular}
    \vspace{0.05in}
    \caption{ This table lists the 50 tasks in \suitename, along with the \pg trajectory sources used to parameterize them. In addition, we list the success and error metrics at the end of training (achieved by our \fname method) for each task, averaged across 3 random seeds. }
    \label{tab:tasks}
\end{table*}

\end{document}